\documentclass{article}
\usepackage{spconf,amsmath,graphicx}
\usepackage{multirow}
\usepackage{color}
\usepackage{makecell}

\usepackage{lipsum}

\let\OLDthebibliography\thebibliography
\renewcommand\thebibliography[1]{
  \OLDthebibliography{#1}
  \setlength{\parskip}{0pt}
  \setlength{\itemsep}{0pt plus 0.5ex}
}

\newcommand*{\sigmoid}{\operatorname{sigmoid}}

\title{Phonetic-and-Semantic Embedding of Spoken Words with Applications in Spoken Content Retrieval}
\name{Yi-Chen Chen, Sung-Feng Huang, Chia-Hao Shen, Hung-yi Lee, Lin-shan Lee}
\address{National Taiwan University, Taiwan}

\begin{document}

%
\maketitle
%
\begin{abstract}
Word embedding or Word2Vec has been successful in offering semantics for text words learned from the context of words.
Audio Word2Vec was shown to offer phonetic structures for spoken words (signal segments for words) learned from signals within spoken words.
This paper proposes a two-stage framework to perform phonetic-and-semantic embedding on spoken words considering the context of the spoken words.
Stage 1 performs phonetic embedding with speaker characteristics disentangled. 
Stage 2 then performs semantic embedding in addition. 
We further propose to evaluate the phonetic-and-semantic nature of the audio embeddings obtained in Stage 2 by parallelizing with text embeddings.

In general, phonetic structure and semantics inevitably disturb each other.
For example the words ``brother" and ``sister" are close in semantics but very different in phonetic structure, while the words ``brother" and ``bother" are in the other way around.
But phonetic-and-semantic embedding is attractive, as shown in the initial experiments on spoken document retrieval.
Not only spoken documents including the spoken query can be retrieved based on the phonetic structures, but spoken documents semantically related to the query but not including the query can also be retrieved based on the semantics.
\end{abstract}

%

\section{Introduction}
\label{sec:intro}

Word embedding or Word2Vec~\cite{mikolov2013distributed,pennington2014glove,bojanowski2016enriching,tu2017learning} has been widely used in the area of natural language processing~\cite{yu2017character,lample2016neural,plank2016multilingual,kim2016character,ballesteros2015improved,DBLP:journals/corr/LuongPM15,DBLP:journals/corr/BahdanauCB14}, in which text words are transformed into vector representations of fixed dimensionality~\cite{sutskever2014sequence,DBLP:conf/iccv/VenugopalanRDMD15,DBLP:journals/corr/KonstasIYCZ17}.
This is because these vector representations carry plenty of semantic information learned from the context of the considered words in the text training corpus.
Similarly, audio Word2Vec has also been proposed in the area of speech signal processing, in which spoken words (signal segments for words without knowing the underlying word it represents) are transformed into vector representations of fixed dimensionality~\cite{he2016multi,settle2016discriminative,chung2016audio,kamper2016deep,bengio2014word,levin2013fixed,toshniwal2017multitask,settle2017query,cho2014learning,DBLP:journals/corr/abs-1803-08976,chen2018towards}.
These vector representations carry the phonetic structures of the spoken words learned from the signals within the spoken words, and have been shown to be useful in spoken term detection, in which the spoken terms are detected simply based on the phonetic structures.
Such Audio Word2Vec representations do not carry semantics, because they are learned from individual spoken words only without considering the context.

Audio Word2Vec was recently extended to Segmental Audio Word2Vec~\cite{wang2018segmental}, in which an utterance can be automatically segmented into a sequence of spoken words~\cite{tran2017parsing,tang2017end,kamper2017embedded,kamper2017segmental} and then transformed into a sequence of vectors of fixed dimensionality by Audio Word2Vec, and the spoken word segmentation and Audio Word2Vec can be jointly trained from an audio corpus.
In this way the Audio Word2Vec was upgraded from word-level to utterance-level.
This offers the opportunity for Audio Word2Vec to include semantic information in addition to phonetic structures, since the context among spoken words in utterances bring semantic information.
This is the goal of this work, and this paper reports the first set of results towards such a goal.

In principle, the semantics and phonetic structures in words inevitably disturb each other.
For example, the words ``brother" and ``sister" are close in semantics but very different in phonetic structure, while the words ``brother" and ``bother" are close in phonetic structure but very different in semantics.
This implies the goal of embedding both phonetic structures and semantics for spoken words is naturally very challenging.
Text words can be trained and embedded as vectors carrying plenty of semantics because the phonetic structures are not considered at all.
On the other hand, because spoken words are just a different version of representations for text words, it is also natural to believe they do carry some semantic information, except disturbed by phonetic structures plus some other acoustic factors such as speaker characteristics and background noise~\cite{DBLP:journals/corr/KulkarniWKT15,DBLP:journals/corr/abs-1709-07902,NIPS2016_6399,higgins2016beta,DBLP:journals/corr/BousmalisTSKE16,DBLP:journals/corr/abs-1711-08010}.
So the goal of embedding spoken words to carry both phonetic structures and semantics is possible, although definitely hard.

But a nice feature of such embeddings is that they may include both phonetic structures and semantics~\cite{jansen2017unsupervised,jansen2017towards}.
A direct application for such phonetic-and-semantic embedding of spoken words is spoken document retrieval~\cite{lee2015spoken,chen2012spoken,chen2011leveraging,lee2013enhancing,wen2013interactive}.
This task is slightly different from spoken term detection, in the latter case spoken terms are simply detected based on the phonetic structures.
Here the goal of the task is to retrieve all spoken documents (sets of consecutive utterances) relevant to the spoken query, which may or may not include the query.
For example, for the spoken query of ``President Donald Trump", not only those documents including the spoken query should be retrieved based on the phonetic structures, but those documents including semantically related words such as ``White House" and ``trade policy", but not necessarily ``President Donald Trump", should also be retrieved.
This is usually referred to as ``semantic retrieval", which can be achieved by the phonetic-and-semantic embedding discussed here.

This paper proposes a two-stage framework of phonetic-and-semantic embedding for spoken words.
Stage 1 performs phonetic embedding but with speaker characteristics disentangled using separate phonetic and speaker encoders and a speaker discriminator.
Stage 2 then performs semantic embedding in addition.
We further propose to evaluate the phonetic-and-semantic nature of the audio embeddings obtained in Stage 2 by parallelizing with text embeddings~\cite{conneau2017word,hoshen2018iterative}.
Very encouraging results including those for an application task of spoken document retrieval were obtained in the initial experiments\footnote{The code is released at https://github.com/grtzsohalf/Audio-Phonetic-and-Semantic-Embedding.git}.

\section{Proposed Approach}
\label{sec:method}

The proposed framework of phonetic-and-semantic embedding of spoken words consists of two stages:

Stage 1 - Phonetic embedding with speaker characteristics disentangled.

Stage 2 - Semantic embedding over phonetic embeddings obtained in Stage 1.

In addition, we propose an approach for parallelizing the audio and text embeddings to be used for evaluating the phonetic and semantic information carried by the audio embeddings.
These are described in Subsections~\ref{subsec:stage1}, ~\ref{subsec:stage2} and~\ref{subsec:parallelize} respectively.

\subsection{Stage 1 - Phonetic Embedding with Speaker Characteristics Disentangled} \label{subsec:stage1}

\begin{figure}[t]
  \centering
  \includegraphics[width=\linewidth]{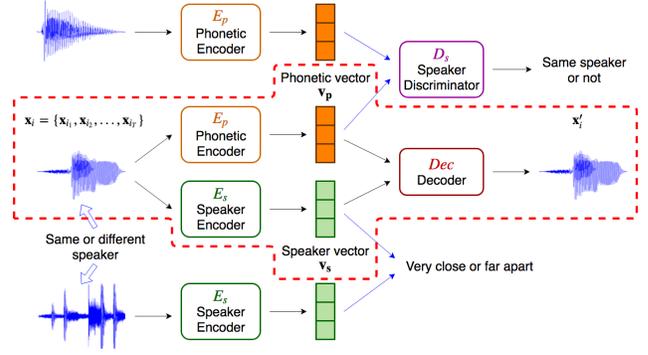}
  \caption{Phonetic embedding with speaker characteristics disentangled.}
  \label{fig:autoencoder}
\end{figure}

A text word with a given phonetic structure corresponds to infinite number of audio signals with varying acoustic factors such as speaker characteristics, microphone characteristics, background noise, etc.
All the latter acoustic factors are jointly referred to as speaker characteristics here for simplicity, which obviously disturbs the goal of phonetic-and-semantic embedding.
So Stage 1 is to obtain phonetic embeddings only with speaker characteristics disentangled.

Also, because the training of phonetic-and-semantic embedding is challenging, in the initial effort we slightly simplify the task by assuming all training utterances have been properly segmented into spoken words.
Because there exist many approaches for segmenting utterances automatically~\cite{wang2018segmental}, and automatic segmentation plus phonetic embedding of spoken words has been successfully trained and reported before~\cite{wang2018segmental}, such an assumption is reasonable here.

We denote the audio corpus as $\mathbf{X} = {\{\mathbf{x}_{i}\}}_{i=1}^{M}$, which consists of $M$ spoken words, each represented as $\mathbf{x}_i=(\mathbf{x}_{i_1}, \mathbf{x}_{i_2}, ..., \mathbf{x}_{i_T})$, where $\mathbf{x}_{i_t}$ is the acoustic feature vector for the t\textsuperscript{th} frame and $T$ is the total number of frames in the spoken word. 
The goal of Stage 1 is to disentangle the phonetic structure and speaker characteristics in acoustic features, and extract a vector representation for the phonetic structure only.

\subsubsection{Autoencoder}
As shown in the middle of Figure~\ref{fig:autoencoder}, a sequence of acoustic features $\mathbf{x}_i=(\mathbf{x}_{i_1}, \mathbf{x}_{i_2}, ..., \mathbf{x}_{i_T})$ is entered to a phonetic encoder $E_p$ and a speaker encoder $E_s$ to obtain a phonetic vector $\mathbf{v_p}$ in orange and a speaker vector $\mathbf{v_s}$ in green. 
Then the phonetic and speaker vectors $\mathbf{v_p}$, $\mathbf{v_s}$ are used by the decoder $Dec$ to reconstruct the acoustic features $\mathbf{x}'$. 
This phonetic vector $\mathbf{v_p}$ will be used in the next stage as the phonetic embedding.
The two encoders $E_p$, $E_s$ and the decoder $Dec$ are jointly learned by minimizing the reconstruction loss below:
\begin{equation}
\begin{aligned}
L_r &= \sum_{i} \| \mathbf{x}_i - Dec(E_p(\mathbf{x}_i), E_s(\mathbf{x}_i)) \|_2^2. 
  \label{reconstruction_loss}
\end{aligned}
\end{equation} 
It will be clear below how to make $E_p$ and $E_s$ separately encode the phonetic structure and speaker characteristics.

\subsubsection{Training Criteria for Speaker Encoder} \label{subsubsec:speaker}
The speaker encoder training requires speaker information for the spoken words.
Assume the spoken word $\mathbf{x}_i$ is uttered by speaker $s_i$.
When the speaker information is not available, we can simply assume that the spoken words in the same utterance are produced by the same speaker. 
As shown in the lower part of Figure~\ref{fig:autoencoder}, $E_s$ is learned to minimize the following loss:
\begin{equation}
\begin{aligned}
L_s &= \sum_{s_i = s_j} \| \mathbf{v_s}_i - \mathbf{v_s}_j \|_2^2 \\
  &+ \sum_{s_i \neq s_j}  \max( \lambda - \|  \mathbf{v_s}_i - \mathbf{v_s}_j \|_2^2, 0).
  \label{speaker_loss}
\end{aligned}
\end{equation} 
In other words, if $\mathbf{x}_i$ and $\mathbf{x}_j$ are uttered by the same speaker ($s_i = s_j$), we want their speaker embeddings $\mathbf{v_s}_i$ and $\mathbf{v_s}_j$ to be as close as possible.
But if $s_i \neq s_j$, we want the distance between $\mathbf{v_s}_i$ and $\mathbf{v_s}_j$ larger than a threshold $\lambda$.

\subsubsection{Training Criteria for Phonetic Encoder} \label{subsubsec:phonetic}

As shown in the upper right corner of Figure~\ref{fig:autoencoder}, a speaker discriminator $D_s$ takes two phonetic vectors $\mathbf{v_p}_i$ and $\mathbf{v_p}_j$ as input and tries to tell if the two vectors come from the same speaker. 
The learning target of the phonetic encoder $E_p$ is to "fool" this speaker discriminator $D_s$, keeping it from discriminating the speaker identity correctly. 
In this way, only the phonetic structure information is learned in the phonetic vector $\mathbf{v_p}$, while only the speaker characteristics is encoded in the speaker vector $\mathbf{v_s}$. 
The speaker discriminator $D_s$ learns to maximize $L_d$ in (\ref{discriminative_loss}), while the phonetic encoder $E_p$ learns to minimize $L_d$, 
\begin{equation}
\begin{aligned}
L_d &= \sum_{s_i = s_j} D_s(\mathbf{v_p}_i, \mathbf{v_p}_j) - \sum_{s_i \neq s_j} D_s(\mathbf{v_p}_i, \mathbf{v_p}_j).
  \label{discriminative_loss}
\end{aligned}
\end{equation} 
where $D_s(\cdot, \cdot)$ is a real number.

\subsubsection{Overall Optimization of Stage 1}
The optimization procedure of Stage 1 consists of four parts: (1) training $E_p$, $E_s$ and $Dec$ by minimizing $L_r$, (2) training $E_s$ by minimizing $L_s$, (3) training $E_p$ by minimizing $L_d$, and (4) training $D_s$ by maximizing $L_d$.
Parts (1)(2)(3) are jointly trained together, while iteratively trained with part (4)~\cite{gulrajani2017improved}.

\subsection{Stage 2 - Semantic Embedding over Phonetic Embeddings Obtained in Stage 1} \label{subsec:stage2}

\begin{figure}[t]
  \centering
  \includegraphics[width=\linewidth]{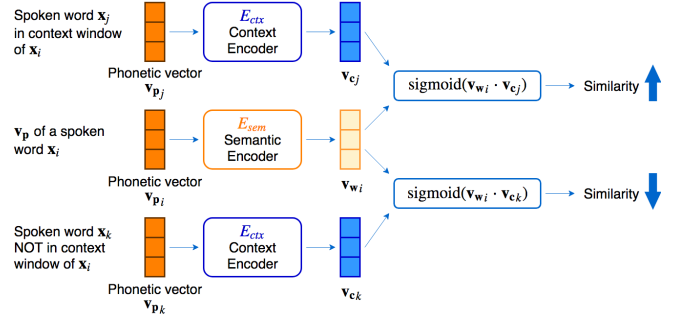}
  \caption{Semantic embedding over phonetic embeddings obtained in Stage 1.}
  \label{fig:semantic}
\end{figure}

As shown in Figure~\ref{fig:semantic}, similar to the Word2Vec skip-gram model~\cite{mikolov2013distributed}, we use two encoders: semantic encoder $E_{\text{sem}}$ and context encoder $E_{\text{ctx}}$ to embed the semantics over phonetic embeddings $\mathbf{v_p}$ obtained in Stage 1.
On the one hand, given a spoken word $\mathbf{x_i}$, we feed its phonetic vector $\mathbf{v_p}_i$ obtained from Stage 1 into $E_{\text{sem}}$ as in the middle of Figure~\ref{fig:semantic}, producing the semantic embedding (in yellow) of the spoken word $\mathbf{v_w}_i = E_{\text{sem}}(\mathbf{v_p}_i) $. 
On the other hand, given the context window size $c$, which is a hyperparameter, if a spoken word $\mathbf{x_j}$ is in the context window of $\mathbf{x_i}$, then its phonetic vector $\mathbf{v_p}_j$ is a context vector of $\mathbf{v_p}_i$. 
For each context vector $\mathbf{v_p}_j$ of $\mathbf{v_p}_i$, we feed it into the context encoder $E_{\text{ctx}}$ in the upper part of Figure~\ref{fig:semantic}, and the output is the context embedding $\mathbf{v_c}_j = E_{\text{ctx}}(\mathbf{v_p}_j)$.

Given a pair of phonetic vectors $(\mathbf{v_p}_i, \mathbf{v_p}_j)$, the training criteria for $E_{\text{sem}}$ and $E_{\text{ctx}}$ is to maximize the similarity between $\mathbf{v_w}_i$ and $\mathbf{v_c}_j$ if $\mathbf{v_p}_i$ and $\mathbf{v_p}_j$ are contextual, while minimizing the similarity otherwise.
The basic idea is parallel to that of text Word2Vec.
Two different spoken words having similar context should have similar semantics. Thus if two different phonetic embeddings corresponding to two different spoken words have very similar context, they should be close to each other after projected by the semantic encoder $E_{\text{sem}}$. 
The semantic and context encoders $E_{\text{sem}}$ and $E_{\text{ctx}}$ learn to minimize the semantic loss $L_{\text{sem}}$ as follows: 
\begin{equation}
\begin{aligned}
L_{\text{sem}} &=  \sum_{(\mathbf{x_i}, \mathbf{x_j}) \text{~in context window}} - \log(\sigmoid(\mathbf{v_w}_i \cdot \mathbf{v_c}_j)) \\
  &+ \sum_{(\mathbf{x_i}, \mathbf{x_k}) \text{~not in context window}} - \log(\sigmoid(- \mathbf{v_w}_i \cdot \mathbf{v_c}_k)). 
  \label{semantic_loss}
\end{aligned}
\end{equation} 
The sigmoid of dot product of $\mathbf{v_w}$ and $\mathbf{v_c}$ is used to evaluate the similarity.
With (\ref{semantic_loss}), if $\mathbf{x_i}$ and $\mathbf{x_j}$ are in the same context window, we want $\mathbf{v_w}_i$ and $\mathbf{v_c}_j$ to be as similar as possible. 
We also use the negative sampling technique, in which only some pairs $(\mathbf{x_i},\mathbf{x_k})$ are randomly sampled as negative examples instead of enumerating all possible negative pairs.

\subsection{Parallelizing Audio and Text Embeddings for Evaluation Purposes}
\label{subsec:parallelize}

In this paper we further propose an approach of parallelizing a set of audio embeddings (for spoken words) with a set of text embeddings (for text words) which will be useful in evaluating the phonetic and semantic information carried by these embeddings.

Assume we have the audio embeddings for a set of spoken words $\mathbf{P_W} = $
$\{\mathbf{p_w}_1, ...,\mathbf{p_w}_i, ..., \mathbf{p_w}_M\}$, where $\mathbf{p_w}_i$ is the embedding obtained for a spoken word $\mathbf{x}_i$ and $M$ is the total number of distinct spoken words in the audio corpus. 
On the other hand, assume we have the text embeddings $\mathbf{Q_W} = $ $\{\mathbf{q_w}_1, ...,\mathbf{q_w}_j, ...,\mathbf{q_w}_M\}$, where $\mathbf{q_w}_j$ is the embedding of the $j$-th text word for the $M$ distinct text words.
Although the distributions of $\mathbf{P_W}$ and $\mathbf{Q_W}$ in their respective spaces are not parallel, that is, a specific dimension in the space for $\mathbf{p_w}$ does not necessarily correspond to a specific dimension in the space for $\mathbf{q_w}$, there should exist some consistent relationship between the two distributions.
For example, the relationships among the words \{France, Paris, Germany\} learned from context should be consistent in some way, regardless of whether they are in text or spoken form.
So we try to learn a mapping relation between the two spaces.
It will be clear below such a mapping relation can be used to evaluate the phonetic and semantic information carried by the audio embeddings.

Mini-Batch Cycle Iterative Closest Point (MBC-ICP)~\cite{hoshen2018iterative} previously proposed as described below is used here. 
Given two sets of embeddings as mentioned above, $\mathbf{P_W}$ and $\mathbf{Q_W}$, they are first projected to their respective top $K$ principal components by PCA. 
Let the projected sets of vectors of $\mathbf{P_W}$ and $\mathbf{Q_W}$ be $\mathbf{A}$ and $\mathbf{B}$ respectively.
If $\mathbf{P_W}$ can be mapped to the space of $\mathbf{Q_W}$ by an affine transformation, the distributions of $\mathbf{A}$ and $\mathbf{B}$ would be similar after PCA~\cite{hoshen2018iterative}.

Then a pair of transformation matrices, $\mathbf{T_{ab}}$ and $\mathbf{T_{ba}}$, is learned, where $\mathbf{T_{ab}}$ transforms a vector $\mathbf{a}$ in $\mathbf{A}$ to the space of $\mathbf{B}$, that is, $\tilde{\mathbf{b}} = \mathbf{T_{ab}}\mathbf{a}$, while $\mathbf{T_{ba}}$ maps a vector $\mathbf{b}$ in $\mathbf{B}$ to the space of $\mathbf{A}$.
$\mathbf{T_{ab}}$ and $\mathbf{T_{ba}}$ are learned iteratively by the algorithm proposed previously~\cite{hoshen2018iterative}.

In our evaluation as mentioned below, labeled pairs of the audio and text embeddings of each word is available, that is, we know $\mathbf{a_i}$ and $\mathbf{b_i}$ for each word $\mathbf{w_i}$. 
So we can train the transformation matrices $\mathbf{T_{ab}}$ and $\mathbf{T_{ba}}$ using the gradient descent method to minimize the following objective function:
\begin{equation}
\begin{aligned}
L_{trans} = & \sum_{i} \|\mathbf{b_i} - \mathbf{T_{ab}}\mathbf{a_i} \|_2^2 +  \sum_{j} \|\mathbf{a_j} - \mathbf{T_{ba}}\mathbf{b_j} \|_2^2  \\
&+ \lambda^\prime \sum_{i} \| \mathbf{a_i} - \mathbf{T_{ba}}\mathbf{T_{ab}}\mathbf{a_i} \|_2^2 \\
&+ \lambda^\prime \sum_{j} \| \mathbf{b_j} - \mathbf{T_{ab}}\mathbf{T_{ba}}\mathbf{b_j} \|_2^2 .
\end{aligned} \label{eq:align}
\end{equation} 
where the last two terms in (\ref{eq:align}) are cycle-constraints to ensure that both $\mathbf{a_i}$ and $\mathbf{b_j}$ are almost unchanged after transformed to the other space and back.
In this way we say the two sets of embeddings are parallelized.

\section{Experimental Setup}
\label{sec:setup}

\subsection{Dataset}
\label{subsec:dataset}

We used LibriSpeech~\cite{panayotov2015librispeech} as the audio corpus in the experiments, which is a corpus of read speech in English derived from audiobooks. This corpus contains 1000 hours of speech sampled at 16 kHz uttered by 2484 speakers.
We used the ``clean" and ``others" sets with a total of 960 hours, and extracted 39-dim MFCCs as the acoustic features. 

\subsection{Model Implementation}
\label{subsec:implementation}

In Stage 1, The phonetic encoder $E_p$, speaker encoder $E_s$ and decoder $Dec$ were all 2-layer GRUs with hidden layer size 128, 128 and 256, respectively.
The speaker discriminator $D_s$ is a fully-connected feedforward network with 2 hidden layers with size 128.
The value of $\lambda$ we used in $L_s$ in (\ref{speaker_loss}) was set to 0.01.

In Stage 2, the two encoders $E_{sem}$ and $E_{ctx}$ were both 2-hidden-layer fully-connected feedforward networks with size 256. 
The size of embedding vectors was set to be 128.
The context window size was 5, and the negative sampling number was 5.

For parallelizing the text and audio embeddings in Subsection~\ref{subsec:parallelize}, we projected the embeddings to the top 100 principle components, so the affine transformation matrices were $100 \times 100$. 
The mini-batch size was 200, and $\lambda^\prime$ in (\ref{eq:align}) was set to 0.5.

\begin{table}[t]
\scriptsize
\centering
\caption{Top-1 nearest accuracies when parallelizing the different versions of audio and text embeddings for different numbers of pairs of spoken and text words.}
\label{table:top1}
\begin{tabular}{|c|c|c|c|c|}
\hline

\multicolumn{1}{|c|}{} &
\multicolumn{1}{|c|}{} &
\multicolumn{1}{c|}{(a)TXT-ph} &
\multicolumn{1}{c|}{(b)TXT-(se,1h)} &
\multicolumn{1}{c|}{(c)TXT-(se,ph)} \\ \hline \hline

\multirow{3}{*}{\makecell{1000 \\ pairs}} 
 & (i)AUD-ph & \textcolor{red}{\bf{0.637}} & 0.124 & 0.550 \\
 & (ii)AUD-(ph\textsuperscript{-}+se) & 0.519 & 0.322 & \bf{0.750} \\
 & (iii)AUD-(ph+se) & 0.598 & \textcolor{red}{0.339} & \textcolor{red}{\bf{0.800}} \\ \hline \hline
 
\multirow{3}{*}{\makecell{3000 \\ pairs}}
 & (i)AUD-ph & \textcolor{red}{\bf{0.465}} & 0.028 & 0.279 \\
 & (ii)AUD-(ph\textsuperscript{-}+se) & \bf{0.330} & 0.032 & 0.254 \\
 & (iii)AUD-(ph+se) & \bf{0.395} & \textcolor{red}{0.033} & \textcolor{red}{0.313} \\ \hline \hline
 
\multirow{3}{*}{\makecell{5000 \\ pairs}}
 & (i)AUD-ph & \textcolor{red}{\bf{0.362}} & 0.012 & 0.190 \\
 & (ii)AUD-(ph\textsuperscript{-}+se) & \bf{0.263} & 0.022 & 0.173 \\
 & (iii)AUD-(ph+se) & \bf{0.315} & \textcolor{red}{0.023} & \textcolor{red}{0.212} \\ \hline

\end{tabular}
\end{table}

\begin{table}[t]
\scriptsize
\centering
\caption{Top-10 nearest accuracies when parallelizing the different versions of audio and text embeddings for different numbers of pairs of spoken and text words.}
\label{table:top10}
\begin{tabular}{|c|c|c|c|c|}
\hline

\multicolumn{1}{|c|}{} &
\multicolumn{1}{|c|}{} &
\multicolumn{1}{c|}{(a)TXT-ph} &
\multicolumn{1}{c|}{(b)TXT-(se,1h)} &
\multicolumn{1}{c|}{(c)TXT-(se,ph)} \\ \hline \hline

\multirow{3}{*}{\makecell{1000 \\ pairs}}
 & (i)AUD-ph & \textcolor{red}{\bf{0.954}} & 0.355 & 0.898 \\
 & (ii)AUD-(ph\textsuperscript{-}+se) & 0.897 & 0.653 & \bf{0.986} \\
 & (iii)AUD-(ph+se) & 0.945 & \textcolor{red}{0.742} & \textcolor{red}{\bf{0.994}} \\ \hline \hline
 
\multirow{3}{*}{\makecell{3000 \\ pairs}} 
 & (i)AUD-ph & \textcolor{red}{\bf{0.854}} & 0.120 & 0.654 \\
 & (ii)AUD-(ph\textsuperscript{-}+se) & \bf{0.758} & 0.146 & 0.671 \\
 & (iii)AUD-(ph+se) & \bf{0.809} & \textcolor{red}{0.166} & \textcolor{red}{0.752} \\ \hline \hline
 
\multirow{3}{*}{\makecell{5000 \\ pairs}}
 & (i)AUD-ph & \textcolor{red}{\bf{0.774}} & 0.050 & 0.518 \\
 & (ii)AUD-(ph\textsuperscript{-}+se) & \bf{0.658} & 0.109 & 0.544 \\
 & (iii)AUD-(ph+se) & \bf{0.717} & \textcolor{red}{0.111} & \textcolor{red}{0.607} \\ \hline

\end{tabular}
\end{table}

\begin{table*}[t]
\footnotesize
\centering
\caption{Some examples of top-10 nearest neighbors in AUD-(ph+se) (proposed), AUD-ph (with phonetic structure) and TXT-(se,1h) (with semantics). The words in red are the common words of AUD-(ph+se) and AUD-ph, and the words in bold are the common words of AUD-(ph+se) and TXT-(se,1h).}
\label{table:NN}
\begin{tabular}{|c|c|c|c|}
\hline

\multicolumn{1}{|c|}{\textbf{words}} & \multicolumn{1}{c|}{\textbf{AUD-(ph+se)}} & \multicolumn{1}{c|}{\textbf{AUD-ph}} & \multicolumn{1}{c|}{\textbf{TXT-(se,1h)}}
\\ \hline \hline

{\multirow{2}{*}{owned}} 
& \textcolor{red}{own}, \textcolor{red}{only}, unknown, owner, land, & owns, \textcolor{red}{armed}, owen, arm, \textcolor{red}{own}, & visited, introduced, lived, related, \textbf{learned}, \\  & \textcolor{red}{armed}, \textbf{learned}, homes, \textbf{known}, alone & \textcolor{red}{only}, oughtnt, loaned, ode, owing & discovered, met, called, think, \textbf{known} \\  \hline

{\multirow{2}{*}{didn't}} 
& did, sitting, give, \textbf{doesn't}, \textbf{don't}, & giving, \textcolor{red}{bidden}, \textcolor{red}{given}, getting, being, & \textbf{don't}, can't, wouldn't, \textbf{doesn't}, won't, \\  & \textcolor{red}{given}, hadn't, too, \textcolor{red}{bidden}, listen & even, ridden, didnt, deane, givin & i'm, you're, shouldn't, think, want \\  \hline

\end{tabular}
\end{table*}

\begin{table}[t]
\scriptsize
\centering
\caption{Spoken document retrieval performance using two different audio embeddings (AUD-(ph+se) and AUD-ph).}
\label{table:LRAP}
\begin{tabular}{|c|c|c|}
\hline
\multicolumn{1}{|c|}{\textbf{groundtruth}} &
\multicolumn{1}{|c|}{\textbf{AUD-(ph+se)}} & 
\multicolumn{1}{c|}{\textbf{AUD-ph}}
\\ \hline \hline
$D_1$ + $D_2$ & 17.8\%  & 15.6\% 
\\ \hline
$D_2$ & 2.8\% & 1.8\% 
\\ \hline 

\end{tabular}
\end{table}

\begin{table*}[t]
\scriptsize
\centering
\caption{Some retrieval examples of chapters in $D_2$ using AUD-(ph+se) show the advantage of semantics information in phonetic-and-semantic embeddings. The word in red in each row indicates the word with the highest similarity to the query in the chapter.}
\label{table:examples}
\begin{tabular}{|c|c|c|c|c|}
\hline









\multicolumn{1}{|c|}{\textbf{(a) query $q$}} &
\multicolumn{1}{|c|}{\textbf{(b) title of a book $b$}} &
\multicolumn{1}{|c|}{\textbf{(c) chapter}} &
\multicolumn{1}{c|}{\textbf{(d) rank}} &
\multicolumn{1}{c|}{\textbf{(e) the word with the highest similarity to the query}}
\\ \hline \hline

nations & Myths and Legends of All Nations & Prometheus the Friend of Man & 13/5273 & ...and shall marry the \textcolor{red}{king} of that country... \\

Anne & Anne of Green Gables & Mrs. Rachel Lynde Is Surprised & 25/5329 & ...why the worthy \textcolor{red}{woman} finally concluded... \\

German & In a German Pension & Story 13: A Blaze & 22/5232 & ...through the heavy snow towards the \textcolor{red}{town}... \\

castle & Montezuma's Castle and Other Weird Tales & THE STRANGE POWDER... & 3/5141 & ...what is its \textcolor{red}{history} asked doctor Farrington... \\


baron & Surprising Adventures of Baron Munchausen & Chapter 22 & 18/5375 & ...at the \textcolor{red}{palace} and having remained in this situation... \\ \hline


\end{tabular}
\end{table*}

\section{Experimental Results}
\label{sec:exp}

\subsection{Evaluation by Parallelizing Audio and Text Embeddings}
\label{subsec:results}

Each text word corresponds to many audio realizations in spoken form. So we first took the average of the audio embeddings for all those realizations to be the audio embedding for the spoken word considered. In this way, each word has a unique representation in either audio or text form.

We applied three different versions of audio embedding (AUD) on the top 1000, 3000 and 5000 words with the highest frequencies in LibriSpeech: (i) phonetic embedding only obtained in Stage 1 in Subsection~\ref{subsec:stage1} (AUD-ph); (ii) phonetic-and-semantic embedding obtained by Stages 1 and 2 in Subsections~\ref{subsec:stage1},~\ref{subsec:stage2}, except the speaker characteristics not disentangled (AUD-(ph\textsuperscript{-}+se)), or $L_s$, $L_d$ in (\ref{speaker_loss}), (\ref{discriminative_loss}) not considered; (iii) complete phonetic-and-semantic embedding as proposed in this paper including Stages 1 and 2 (AUD-(ph+se)).
So this is for ablation study.

On the other hand, we also obtained three different types of text embedding (TXT) on the same set of top 1000, 3000 and 5000 words.
Type (a) Phonetic Text embedding (TXT-ph) considered precise phonetic structure but not context or semantics at all.
This was achieved by a well-trained sequence-to-sequence autoencoder encoding the precise phoneme sequence of a word into a latent embedding.
Type (b) Semantic Text embedding considered only context or semantics but not phonetic structure at all, and was obtained by a standard skip-gram model using one-hot representations as the input (TXT-(se,1h)).
Type (c) Semantic and Phonetic Text embedding (TXT-(se,ph)) considered context or semantics as well as the precise phonetic structure, obtained by a standard skip-gram model but using the Type (a) Phonetic Text embedding (TXT-ph) as the input.
So these three types of text embeddings provided the reference embeddings obtained from text and/or phoneme sequences, not disturbed by audio signals at all.

Now we can perform the transformation from the above three versions of audio embeddings (AUD-ph, AUD-(ph\textsuperscript{-}+se), AUD-(ph+se)) to the above three types of text embeddings (TXT-ph, TXT-(se,1h), TXT-(se,ph)) by parallelizing the embeddings as described in Subsection~\ref{subsec:parallelize}.
The evaluation metric used for this parallelizing test is the top-k nearest accuracy.
If the audio embedding representation $\mathbf{a_i}$ of a word $\mathbf{w_i}$ is transformed to the text embedding $\mathbf{b_j}$ by $\mathbf{T_{ab}}$, and $\mathbf{b_j}$ is among the top-k nearest neighbors of the text embedding representation $\mathbf{b_i}$ of the same word, this transformation for word $\mathbf{w_i}$ is top-k-accurate.
The top-k nearest accuracy is then the percentage of the words considered which are top-k-accurate.

The results of top-k nearest accuracies for k=1 and 10 are respectively listed in Tables~\ref{table:top1} and~\ref{table:top10}, each for 1000, 3000 and 5000 pairs of spoken and text words.

First look at the top part of Table~\ref{table:top1} for top-1 nearest accuracies for 1000 pairs of audio and text embeddings.
Since column (a) (TXT-ph) considered precise phonetic structures but not semantics at all, the relatively high accuracies in column (a) for all three versions of audio embedding (i)(ii)(iii) implied the three versions of audio embedding were all rich of phonetic information.
But when the semantics were embedded in (ii)(iii) (AUD-(ph\textsuperscript{-}+se), AUD-(ph+se)), the phonetic structures were inevitably disturbed (0.519, 0.598 vs 0.637).
On the other hand, column (b) (TXT-(se,1h)) considered only semantics but not phonetic structure at all, the relatively lower accuracies implied the three versions of audio embedding did bring some good extent of semantics, except (i) AUD-ph, but obviously weaker than the phonetic information in column (a).
Also, the Stage 2 training in rows (ii)(iii) (AUD-(ph\textsuperscript{-}+se), AUD-(ph+se)) gave higher accuracies than row (i) (AUD-ph) (0.339, 0.332 vs 0.124 in column (b)), which implied the Stage 2 training was successful.
However, column (c) (TXT-(se,ph)) is for the text embedding considering both the semantic and phonetic information, so the two versions of phonetic-and-semantic audio embedding for rows (ii)(iii) had very close distributions (0.750, 0.800 in column (c)), or carried good extent of both semantics and phonetic structure.
The above are made clearer by the numbers in bold which are the highest for each row, and the numbers in red which are the highest for each column.
It is also clear that the speaker characteristics disentanglement is helpful, since row (iii) for AUD-(ph+se) was always better than row (ii) for AUD-(ph\textsuperscript{-}+se).

Similar trends can be observed in the other parts of Table~\ref{table:top1} for 3000 and 5000 pairs, except the accuracies were lower, probably because for more pairs the parallelizing transformation became more difficult and less accurate.
The only difference is that in these parts column (a) for TXT-ph had the highest accuracies, probably because the goal of semantic embedding for rows (ii)(iii) (AUD-(ph\textsuperscript{-}+se), AUD-(ph+se)) was really difficult, and disturbed or even dominated by phonetic structures.
Similar trends can be observed in Table~\ref{table:top10} for top-10 accuracies, obviously with higher numbers for top-10 as compared to those for top-1 in Table~\ref{table:top1}.

In Table~\ref{table:NN}, we list some examples of top-10 nearest neighbors in AUD-(ph+se) (proposed), AUD-ph (with phonetic structure) and TXT-(se,1h) (with semantics).
The words in red are the common words for AUD-(ph+se) and AUD-ph, and the words in bold are the common words of AUD-(ph+se) and TXT-(se,1h).
For example, the word ``owned" has two common semantically related words ``learned" and ``known" in the top-10 nearest neighbors of AUD-(ph+se) and TXT-(se,1h).
The word ``owned" also has three common phonetically similar words ``armed", ``own" and ``only" in the top-10 nearest neighbors of AUD-(ph+se) and AUD-ph.
This is even clearer for the function word ``didn't".
These clearly illustrate the phonetic-and-semantic nature of AUD-(ph+se).

\subsection{Results of Spoken Document Retrieval}

The goal here is to retrieve not only those spoken documents including the spoken query (e.g. ``President Donald Trump") based on the phonetic structures, but those including words semantically related to the query word (e.g. ``White House").
Below we show the effectiveness of the phonetic-and-semantc embedding proposed here in this application.

We used the 960 hours of ``clean" and ``other" parts of LibriSpeech dataset as the target archive for retrieval, which consisted of 1478 audio books with 5466 chapters. 
Each chapter included 1 to 204 utterances or 5 to 6529 spoken words.
In our experiments, the queries were the keywords in the book titles, and the spoken documents were the chapters.
We chose 100 queries out of 100 randomly selected book titles, and our goal was to retrieve query-relevant documents.
For each query $q$, we defined two sets of query-relevant documents: The first set $D_1^q$ consisted of chapters which included the query $q$.
The second set $D_2^q$ consisted of chapters whose content didn't contain $q$, but these chapters belonged to books whose titles contain $q$ (so we assume these chapters are semantically related to $q$).
Obviously $D_1^q$ and $D_2^q$ were mutually exclusive, and $D_2^q$ were the target for semantic retrieval, but couldn't be retrieved based on the phonetic structures only.

For each query $q$ and each document $d$, the relevance score of $d$ with respect to $q$, $s(q, d)$, is defined as follows:
\begin{equation}
\begin{aligned}
s(q, d) &= \max_{w\ \text{in}\ d} -\|R(w) - R(q)\|_2,
  \label{score}
\end{aligned}
\end{equation}
where $R(w)$ is the audio embedding of a word $w$ in $d$.
So (\ref{score}) indicates the documents $d$ were ranked by the minimum distance between a word $w$ in $d$ and the query $q$.
We used mean average precision (MAP) as the evaluation metric for the spoken document retrieval test.

We compared the retrieval results with two versions of audio embedding: AUD-(ph+se) and AUD-ph.
The results are listed in Table~\ref{table:LRAP} for two definitions of groundtruth for the query-relevant documents: the union of $D_1$ and $D_2$ and $D_2$ alone.
As can be found from this table, AUD-(ph+se) offered better retrieval performance than AUD-ph in both rows.
Note that those chapters in $D_2$ in the second row of the table did not include the query $q$, so couldn't be well retrieved using phonetic embedding alone. 
That is why the phonetic-and-semantic embedding proposed here can help.

In Table~\ref{table:examples}, we list some chapters in $D_2$ retrieved using AUD-(ph+se) embeddings to illustrate the advantage of the phonetic-and-semantic embeddings.
In this table, column (a) is the query $q$, column (b) is the title of a book $b$ which had chapters in $D_2^q$, column (c) is a certain chapter $chp$ in $b$, column (d) is the rank of $chp$ out of all chapters whose content didn't contain $q$, and column (e) is a part of the content in $chp$ where the word in red is the word in $chp$ with the highest similarity to $q$.
For example, in the first row for the query ``nations", the chapter ``Prometheus the Friend of Man" of the book titled ``Myths and Legends of All Nations" is in $D_2^{nations}$.
The word ``nations" is not in the content of this chapter.
However, because the word ``king" semantically related to ``nations" is in the content, this chapter was ranked the 13\textsuperscript{th} among all chapters whose content didn't contain the word ``nations".
This clearly verified why the semantics in the phonetic-and-semantic embeddings can remarkably improve the performance of spoken content retrieval.

\section{Conclusions and Future Work}
\label{sec:conclusion}

In this paper we propose a framework to embed spoken words into vector representations carrying both the phonetic structure and semantics of the word.
This is intrinsically challenging because the phonetic structure and the semantics of spoken words inevitably disturbs each other.
But this phonetic-and-semantic embedding nature is desired and attractive, for example in the application task of spoken document retrieval.
A parallelizing transformation between the audio and text embeddings is also proposed to evaluate whether such a goal is achieved.


\bibliographystyle{IEEEbib}

\end{document}